\pdfoutput=1

\documentclass[11pt]{article}

\usepackage{acl}
\usepackage{times}
\usepackage{algorithm}
\usepackage{latexsym}
\usepackage{array}
\usepackage{booktabs}
\usepackage{stfloats}
\usepackage{algpseudocode}
\usepackage{float}
\usepackage[T1]{fontenc}

\usepackage[utf8]{inputenc}

\usepackage{multirow}
\usepackage{amsmath}
\usepackage{adjustbox}
\usepackage{makecell}
\usepackage{graphicx}
\usepackage{caption}
\usepackage{subcaption}
\usepackage{xcolor}
\usepackage{microtype}

\title{Systematic Analysis for Pretrained Language Model Priming for Parameter-Efficient Fine-tuning}

\author{Shih-Cheng Huang$^{1*}$, Shih-Heng Wang$^{1*}$, Min-Han Shih$^{2*}$, Saurav Sahay$^{2+}$, and Hung-yi Lee$^{1*}$\\
$^{*}$National Taiwan University, Taipei, Taiwan \\
$^{+}$Intel Labs, Santa Clara, CA, USA \\
\textit{\{r09942093,r11942079,b08502141,hungyilee\}@ntu.edu.tw}$^{*}$\\
\textit{saurav.sahay@intel.com}$^{+}$}
\usepackage{siunitx}  
\begin{document}
\maketitle
\begin{abstract}
    Parameter-efficient (PE) methods (like Prompts or Adapters) for adapting pre-trained language models (PLM) to downstream tasks have been popular recently. However, hindrances still prevent these methods from reaching their full potential. For example, two significant challenges are few-shot adaptation and cross-task generalization. To tackle these issues, we propose a general PE \textbf{priming} framework to enhance and explore the few-shot adaptation and generalization ability of PE methods. In this framework, PLMs are primed with PE methods for rapidly adapting to various target tasks. To evaluate the generalization ability of these PE methods, we conduct experiments on a few-shot cross-domain benchmark containing 160 diverse NLP tasks. Our experiment not only reveals the best priming strategy but also verifies that priming facilitates the adaptation to target tasks.
\end{abstract}

\section{Introduction}
     In recent years, pre-trained language models (PLMs) in natural language processing (NLP) are blooming everywhere \citep{BERT, BART, roberta, joshi2020spanbert, T5, GPT2, GPT3}. However, not only the number of PLMs but also their size is rapidly growing, making it harder to perform full fine-tuning. To address the issue, tons of parameter-efficient fine-tuning (PEFT) methods have bubbled up, such as adapters \citep{adapter-houlsby, pfeiffer2020adapterfusion, zaken2021bitfit, fu2022adapterbias},  or prompts \citep{google_prompt_tuning, prefix_prompt_v1}. These methods have made it equitable for researchers with insufficient computational resources. 
    \begin{figure}[t!]
        \centering
        \includegraphics[width=0.8\linewidth]{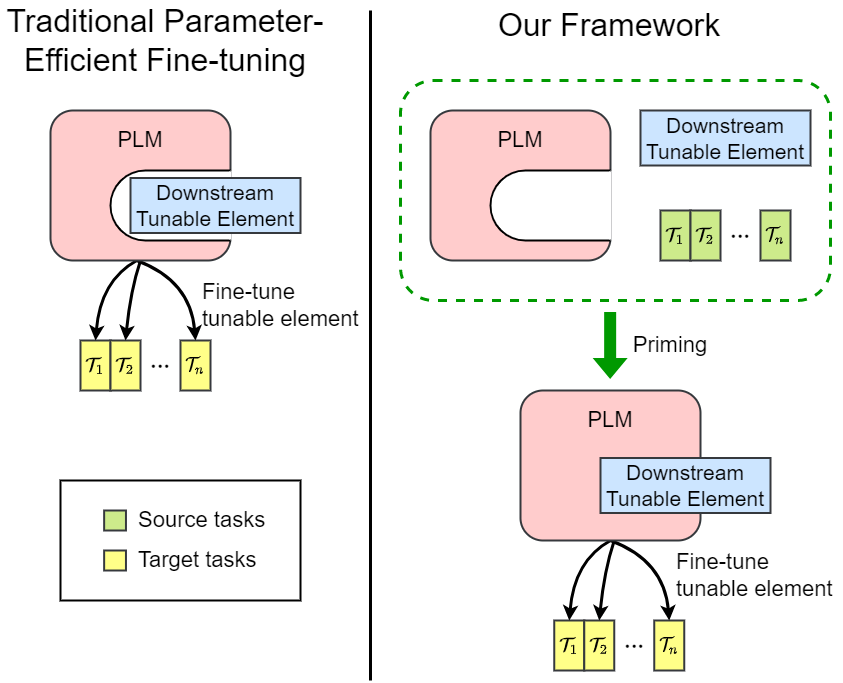}
        \caption{We propose a general framework to improve the performance of parameter-efficient fine-tuning. We prime the PLM with source tasks for parameter-efficient methods. }
        \label{fig:concept}
    \end{figure}
However, there is still a long way to go for these PE methods to reach their full potential. Because the pre-training objectives are not directly related to PE, it is foreseeable that there is a mismatch between the PLM and PE methods, which may prevent PE methods from unleashing their full power. To address this problem, we introduce an additional "priming" stage between pre-training and downstream fine-tuning. As shown in Fig.\ref{fig:concept}, we prime the PLM on extra few-shot source tasks with multitask learning (MTL) or meta-learning, and then fine-tune PE elements to target tasks. Compared with traditional PEFT methods, the PLM and PE elements can fit each other better after priming. In other words, instead of the conventional PEFT paradigm (pre-train $\rightarrow$ PEFT), we adopt the "pre-train $\rightarrow$ priming $\rightarrow$ PEFT" pipeline.

  Some recent studies explore how to bridge the gap between pre-training tasks and target tasks. \citet{PPT} pre-trains the soft prompt tokens with self-supervised tasks to give a better initialization. \citet{LBI,hou-etal-2022-metaprompting} exploits optimization-based meta-learning to find an initialization for soft prompts to facilitate faster adaptation to new tasks. \citet{gheini2022know} tweaks the meta-learning algorithm MAML \citep{MAML} for priming and simulates the PEFT procedure in the inner loop. However, they only focus on priming for a single PE method with a specific algorithm. In contrast, our work views priming as a general method to boost PEFT from a higher perspective.  Moreover, previous works explore only single-domain tasks, which lack the exploration of generalization ability. On the contrary, our work evaluate PE methods with diverse NLP tasks in various domains.

    On top of that, we conduct comprehensive experiments over well-known PE methods like adapters and prompt tuning under different settings. The experiment result reveals that priming by tuning only PLM leads to the best adaptation result on  target tasks. In addition, we shows priming does help the whole model to converge more easily, which validates the necessity of priming.

    
    

\section{Related Work}
    \subsection{Adapter}
    Adapters \cite{adapter-houlsby,PALs,K-adapter,meta-adapter, fu2022adapterbias, pfeiffer2020adapterfusion, karimi2021compacter, hu2021lora, zaken2021bitfit, he2021towards} are lightweight modules introduced for the transformer architecture. Adapters add extra trainable parameters and freeze the original PLM parameters during fine-tuning. 
    
    \subsection{Prompt}
    Prompt-based tuning \cite{prefix_prompt_v1, google_prompt_tuning, soft-prompt, LBI, PPT, p-tuningv2, hard-prompt, hou-etal-2022-metaprompting, Vu2021SPoTBF, Liu2022PTuningPT} is an innovative method to use the power of PLMs efficiently. \citet{PPT} proposed the concept of prompt initialization pre-training. \citet{LBI} proposed Meta-learned Prompt Tuning (MetaPT) to further improve prompt initialization.
    
    

\section{Methodology}
    \label{sect:method}
    \subsection{Framework}
    Our work aims to comprehensively analyze the priming strategies by comparing the performance of PE methods under few-shot scenarios. We introduce a general framework to prime the whole model (may include PLMs) to better adapt to downstream tasks. Our training approach consists of two distinct stages: the \textbf{upstream priming stage} and the \textbf{downstream fine-tuning stage}. Initially, the model acquires knowledge from source tasks during the upstream priming stage, followed by few-shot fine-tuning on target tasks in the downstream stage.
    
    Specifically, we name parameters fine-tuned in the upstream stage as \textbf{upstream tunable elements}, while those in the downstream stage as \textbf{downstream tunable elements}. \textbf{Upstream tunable elements} and \textbf{downstream tunable elements} may be fully-overlapping, partially-overlapping or non-overlapping.  
    
    \subsection{Upstream Priming Stage} \label{ssec:upstream-stage}
    The upstream priming stage is designed to prime the model's upstream tunable elements, enabling it to quickly adapt to a range of downstream few-shot tasks. We employ a \textbf{priming algorithm}, predominantly Meta Learning or Multitask Learning (MTL), to update the upstream tunable elements on source tasks. Upstream tunable elements comprise \textbf{Pre-trained Language Models (PLM)}, \textbf{adapters}, and \textbf{prompts}. In other words, the combination of upstream tu

    \subsubsection{Multi-task Learning}
    In Multi-task Learning (MTL), multiple tasks are learned concurrently by minimizing their combined loss. This method enhances the model's capability to learn cross-task features and accelerates adaptation. Our implementation of MTL focuses on training the upstream tunable elements during the upstream priming stage. We define $\psi$ as the whole model's parameters, with subsets $\psi_u$ for upstream and $\psi_d$ for downstream tunable elements. The objective is to minimize loss across training tasks while adjusting only $\psi_u$:
    
    \begin{equation}\small
        \psi_u^\prime = \mathop{\arg\min}_{\psi_u} \sum_{\mathcal{T}_i\in \mathcal{T}}\mathcal{L}\left(\psi, \mathcal{T}_i\right)
    \end{equation}
    
    Here, $\mathcal{L}$ is the loss function, and $\mathcal{T}_i$ represents the $i^{th}$ task from the set of source tasks $\mathcal{T}$.
    
    It is important to note that if $\psi_d$ is not included in $\psi_u$, it is initialized but remains unchanged during the upstream priming stage. For example, if PLM is selected as the upstream tunable element and adapters as the downstream element, the adapters are initialized in the upstream stage but only tuned during the downstream fine-tuning phase.
    
    \subsubsection{Meta Learning}
\begin{figure}[h!]
        \centering
        \includegraphics[width=0.9\linewidth]{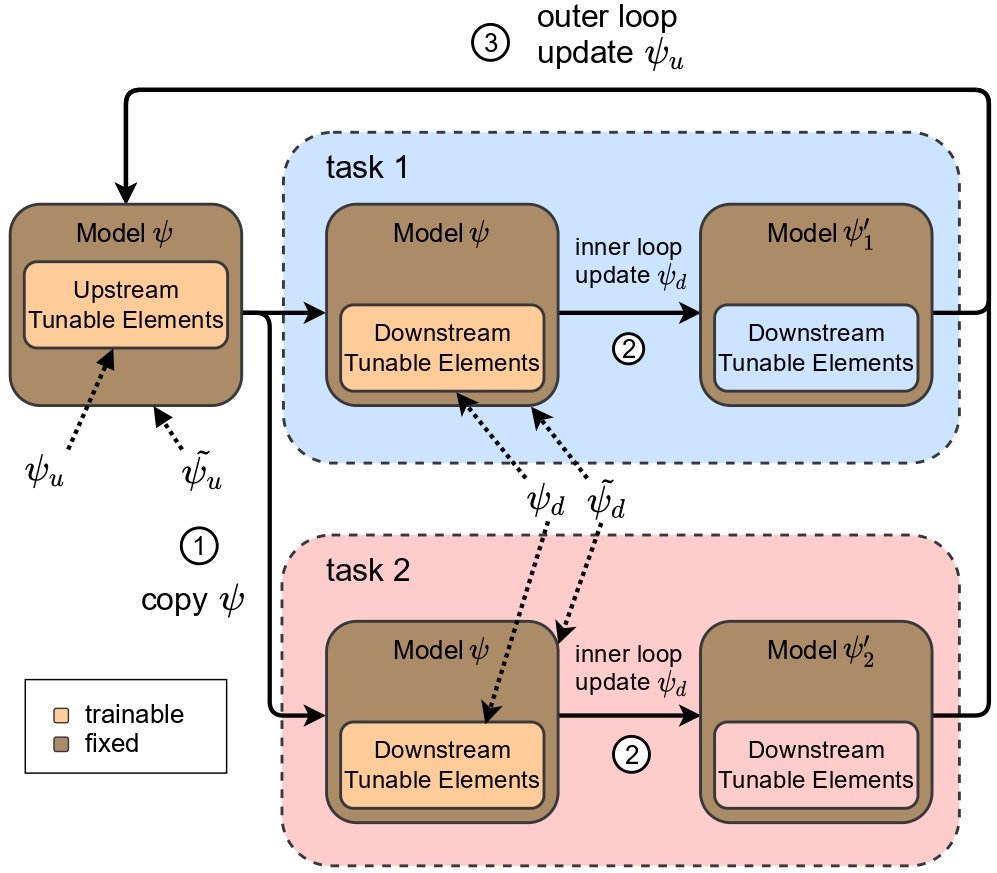}
        \caption{Illustration of Alg.~\ref{alg:cap}}
        \label{fig:maml}
    \end{figure}

In our study, we utilize the MAML\cite{MAML} algorithm, for our priming process. MAML is distinctive in its dual-phase training approach: the inner loop and the outer loop. The inner loop is designed for task-specific adaptation, while the outer loop focuses on finding an optimal initialization for quick adaptation in the inner loop. We modify some parts of MAML algorithm, which are outlined in Alg.\ref{alg:cap} and illustrates in Fig.~\ref{fig:maml}. It starts by copying the current model parameters $\psi$ as the initial state for the inner loop. In this loop, we specifically tune the downstream tunable element $\psi_d$. The tuned parameters for the $i^{th}$ task in the inner loop are represented as $\psi_{i}^{\prime}$. The final step involves calculating the loss from the adapted model $\psi_{i}^{\prime}$ and the source tasks $\mathcal{T}_i$, which is then used to update $\psi_u$. The updated $\psi_u$ are initialized for the subsequent inner loop.
    
    \begin{algorithm}[h]
    \caption{Parameter-Efficient MAML}
    \label{alg:cap}
    \small
    \begin{algorithmic}[1]
    \State  $\mathcal{T}=\{T_1, T_2, ...\}$: A set of source tasks 
    \State  $\alpha, \beta$: Outer lr, Inner lr 
    \State $\theta$: PLM parameters
    \State $\{\phi_1, \phi_2, ...\}$: Tunable elements
    \State $\psi = [\theta;\phi_1;\phi_2;...]$: All parameters of the model
    \\
    \State Randomly initialize $\{\phi_1, \phi_2, ...\}$
    \While{not done}
        \For{$T_i \in \mathcal{T}$}
            \State Split $\psi$ into two parts, $\psi_d$ and $\tilde{\psi_d}$
            \State Evaluate $\nabla_{\boldsymbol{\psi_d}}\mathcal{L}_{T_i}(f_\psi)$ with respect to K samples
            \State Compute adapted parameters with gradient \\\qquad\quad descent: $\psi_{d,i}^{\prime}=\boldsymbol{\psi_d}-\beta \nabla_{\boldsymbol{\psi_d}} \mathcal{L}_{T_i}\left(f_{\psi}\right)$
            \State $\psi_i^\prime = [\psi_{d,i}^{\prime};\tilde{\psi_d}]$
            
        \EndFor
        \State Split $\psi$ into two parts, $\psi_u$ and $\tilde{\psi_u}$
        \State $\psi_u^\prime = \psi_u-\alpha \nabla_{\psi_u} \sum_{\mathcal{T}_i \sim p(\mathcal{T})} \mathcal{L}_{T_i}(f_{\psi_i^\prime})$
        \State $\psi \leftarrow [\psi_u^\prime;\tilde{\psi_u}]$

    \EndWhile
    
    \State \Return $\psi$

    \end{algorithmic}
    \end{algorithm}

    \subsection{Downstream Fine-Tuning Stage} \label{ssed:downstream-stage}
     In downstream fine-tuning stage, with the primed initialization obtained from the upstream priming stage, we directly fine-tune the downstream tunable elements on the target tasks. Since the backbone of our work is to explore the cross-domain few-shot ability of PEFT methods w \& w/o priming, \textbf{only prompt or adapter are tunable in downstream stage}.  
    



\section{Experiment} \label{sec:experiment}
    \subsection{Dataset}
    We choose \textbf{CrossFit Challenge} \cite{crossfit} as our benchmark, which provides 160 different NLP few-shot tasks with a unified text-to-text format gathered from existing open-access datasets. 

    In CrossFit Challenge, they divide all tasks into non-overlapping Train, Dev, and Test tasks. We select \textbf{random split} in \citet{crossfit} to be the task split setting in our work. 
    We select the Train tasks as the source tasks for upstream priming and the Test tasks as the target tasks for downstream fine-tuning. More explicit explanations of tasks can be found in \citet{crossfit}. Briefly speaking, CrossFit Challenge is able to evaluate the authentic few-shot generalization ability of models. 
    
    \subsection{Setup}
    \subsubsection{Tunable Elements}
    Our experiment setup mainly follows \citet{crossfit}. In the upstream priming stage, we can tune prompt, adapter and PLM, but we only tune prompt or adapters during the downstream fine-tuning stage to accord with the spirit of PE methods. It's crucial to emphasize that the initialized parameters obtained from the upstream priming stage are carried forward to the downstream fine-tuning stage. 
    
    \subsubsection*{Adapter} \label{sssec:adapter}
    In this work, we mainly adopt AdapterBias \citep{fu2022adapterbias} as our adapter module. AdapterBias adds a token-dependent shift to the hidden output of transformer layers, parameterized by only a vector and a linear layer. Compared with the original adapter design \citep{adapter-houlsby}, the trainable parameters are further reduced while obtaining comparable performance.
    
    \subsubsection*{Prompt} \label{sssec:prompt}
    Prompt is one of our tunable elements. In our settings, we applied prompt tuning proposed by \citet{google_prompt_tuning}, which concatenates tunable tokens before the input sentence and ask the PLM to generate corresponding output text. Following the settings in \citet{google_prompt_tuning}, we set the prompt length to 100 tokens.
    
    
    
    \subsubsection{Hyperparameters}
    In our research, we use the BART-base model~\cite{wolf2019huggingface} as our primary language model. For both the MTL and the outer loop of meta-learning, we employ the AdamW~\cite{loshchilov2017decoupled} with a weight decay of 0.01. Specifically in meta-learning, we set different outer loop learning rates for various elements: $8\times10^{-5}$ for PLMs, $8\times10^{-3}$ for prompts, and $1\times10^{-5}$ for adapters. The inner loop has its learning rates set at 0.025 for prompts and 0.001 for adapters. The training is conducted over 80 epochs, with a batch size of 1 for training and an inner batch size of either 4 or 8, contingent on GPU memory limits. For MTL, we maintain a consistent learning rate of $3\times10^{-5}$ for PLMs, prompts, and adapters. The MTL training spans 10 epochs with a train batch size of 32.

    \subsection{Metrics}
    For the evaluation metric, we also follow \citet{crossfit}, adopting \emph{Average Relative Gain} (\textbf{ARG}) as one of the performance indexes, and the definition for ARG is: 
    \begin{equation}\small
        \text{ARG} = \frac{1}{n}\sum_{i=1}^{n}(\frac{\text{$P^i$} - \text{$P_0^i$}}{\text{$P_0^i$}})
    \end{equation}
    $\theta$ : Adapter and downstream model
    $P_0^i$ represents the performance of directly fine-tuning PLMs on $i^{th}$ target tasks, and $P_i$ is that of our experiment combination. Since the comparing target is directly fine-tuning PLMs, baselines with ARG greater than 0 are those that surpass fine-tuning PLMs. 
    
\begin{table*}[h!]
    \begin{adjustbox}{max width=1\linewidth}
    \centering
    \begin{tabular}{cl*{8}{c}}
    \Xhline{1.5pt}
     &  & & \textbf{Clf-F1} & \textbf{ACC} & \textbf{EM} & \textbf{Matthew Cor.} & \textbf{QA-F1} & \textbf{Rouge-L} & \textbf{ARG} \\ \cline{4-10}

     \multirow{3}{*}{\makecell{Without\\Priming}}
     & \multirow{3}{*}{\makecell{Direct\\Fine-tuning}}
 &    PLM (M) (100\%)    & \textbf{0.6474} & \textbf{0.5839} & 0.3304           & \textbf{0.0896}  & \textbf{0.2919}    & 0.8033          & \textbf{0.0000} \\ 
 & & Prompt (P) (0.06\%) & 0.5570          & 0.5115          & \textbf{0.4147}  & 0.0495           & 0.2765             & 0.8030          &  -0.1040 \\ 
 & & Adapter (A) (0.07\%) & 0.4503          & 0.4784          & 0.2824           & -0.0276          & 0.2863             & \textbf{0.8081} &  -0.2886 \\ 
         \cline{1-10} 
     \multirow{16}{*}{\makecell{With\\Priming}}
 & \multirow{6}{*}{\makecell{Meta \\ Learning}}
 &   P\_P     & \textbf{0.6283} & 0.5321            & \textbf{0.4291} & 0.0436          & 0.3261          & \textbf{0.8010}  &  -0.0331$^*$ \\
 & & M\_P     & 0.6267          & \textbf{0.6441}   & 0.1961          & 0.0505          & 0.3064          & 0.7962           &  \textbf{-0.0143}$^*$  \\
 & & M+P\_P   & 0.6173          & 0.5783            & 0.1777          & \textbf{0.0626} & 0.2509          & 0.7690           &  -0.0477$^*$ \\
 & & A\_A     & 0.4209          & 0.4257            & 0.2326          & -0.0556         & 0.2899          & 0.7958           &  -0.3534 \\
 & & M\_A     & 0.5546          & 0.6457            & 0.2253          & 0.0067          & 0.2656          & 0.7284           &  -0.1045$^*$ \\
 & & M+A\_A   & 0.3528          & 0.4614            & 0.2580          & -0.0050         & \textbf{0.3868} & 0.7479           &  -0.3025 \\ 
  \cline{2-10} 
 & \multirow{6}{*}{\makecell{Multi-Task \\ Learning}}
 &   P\_P     & 0.5524          & 0.5088            & \textbf{0.4055} & \textbf{0.0765} & 0.3390         & 0.7993            &  -0.0863$^*$ \\
 & & M\_P     & \textbf{0.6646} & 0.6491            & 0.2509          & 0.0612          & 0.3841         & \textbf{0.8086}   &   0.0488$^*$ \\
 & & M+P\_P   & 0.6610          & \textbf{0.6519}   & 0.2622          & 0.0716          & 0.3943         & 0.8032            &   \textbf{0.0571}$^*$ \\
 & & A\_A     & 0.3358          & 0.4274            & 0.2743          & -0.0469         & 0.3007         & 0.7405            &  -0.3808 \\
 & & M\_A     &  0.6122  &   0.6496  &     0.2821    &   -0.0128     & \textbf{0.4432} & 0.7383   & -0.0158$^*$ \\
 & & M+A\_A   & 0.3815          & 0.5709            & 0.2048          & -0.0483         & 0.4081         & 0.6713            & -0.2531$^*$ \\
 \Xhline{1.5pt}

    \end{tabular}
    \end{adjustbox}
    \caption{This table shows the detailed performance of different baselines. We divide the methods by whether it is primed and its priming algorithm. The percentages beside each setting represent the proportion of parameters trained in the downstream fine-tuning stage. } 
    \label{tab:main-experiment-combinations}
\end{table*}
	
   \subsection{Annotation}
       We use abbreviations to make the result more concise, including \textbf{M} for PLM, \textbf{P} for prompts, and \textbf{A} for adapters. Additionally, characters before the underline represent the upstream tunable elements, while those after the underline represent the downstream tunable elements of the baseline. For example, P\_P represents upstream and downstream tunable elements are both prompts; M+A\_A represents upstream tunable elements are PLM and adapters, and downstream tunable element is adapters. Lastly, we use FT\_M, FT\_A, and FT\_P to represent directly fine-tuning the PLM, adapter, and prompt, respectively.  
       
    \subsection{Main Result}

    The first block in Table \ref{tab:main-experiment-combinations} showcases baselines without priming, while subsequent blocks feature combinations of different priming strategies categorized by the priming algorithm. Among the direct fine-tuning results, fine-tuning PLM serves as a competitive baseline with high training costs, while direct fine-tuning prompts/adapters are considered as primary baselines against each priming prompts/adapters baseline.
    
    Table \ref{tab:main-experiment-combinations} underscores the effectiveness of priming, with most baselines showing noticeable improvements (indicated by asterisks$^\ast$) across various priming algorithms. Notably, some combinations outperform direct fine-tuning of prompts, and a few even surpass fine-tuning PLM, such as M\_P and M+P\_P in multi-task learning (ARGs greater than 0). For adapters, certain combinations demonstrate remarkable progress, like M\_A in both meta-learning and multi-task learning, while others experience slight drops in performance, such as A\_A in both meta-learning and multi-task learning.
    \subsection{Parameter Efficiency}
\begin{figure}[h!]
        \centering
        \includegraphics[width=0.83\linewidth]{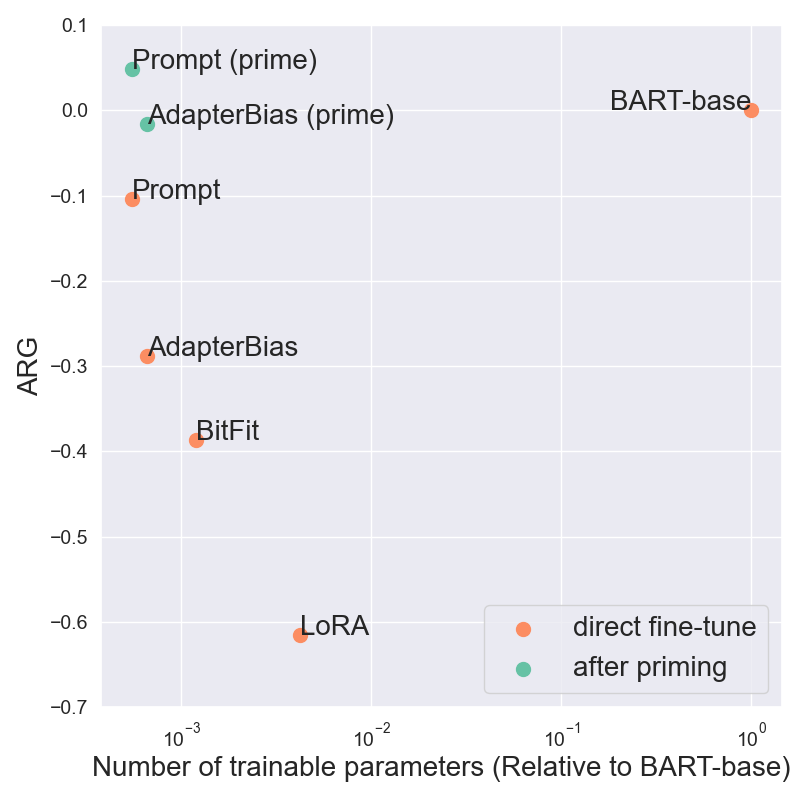}
        \caption{This figure shows the parameter efficiency of different baselines. }
        \label{fig:params}
    \end{figure}

    Fig~\ref{fig:params} visually depicts the relationship between the performance of each method and its tuned parameter scale. The best-performing combinations from Table \ref{tab:main-experiment-combinations} represent priming prompts/adapters (green ones). Other baselines include fine-tuning prompts/adapters without priming, fine-tuning PLM (BART-base), and existing works like LoRA\cite{hu2021lora} and BitFit\cite{zaken2021bitfit}. Fig~\ref{fig:params} demonstrates that priming prompts/adapters baselines locate at the upper-left region, indicating superior results and higher parameter efficiency brought by priming.

\section{Analysis}

    \begin{figure}[t]
        \includegraphics[width=\linewidth]{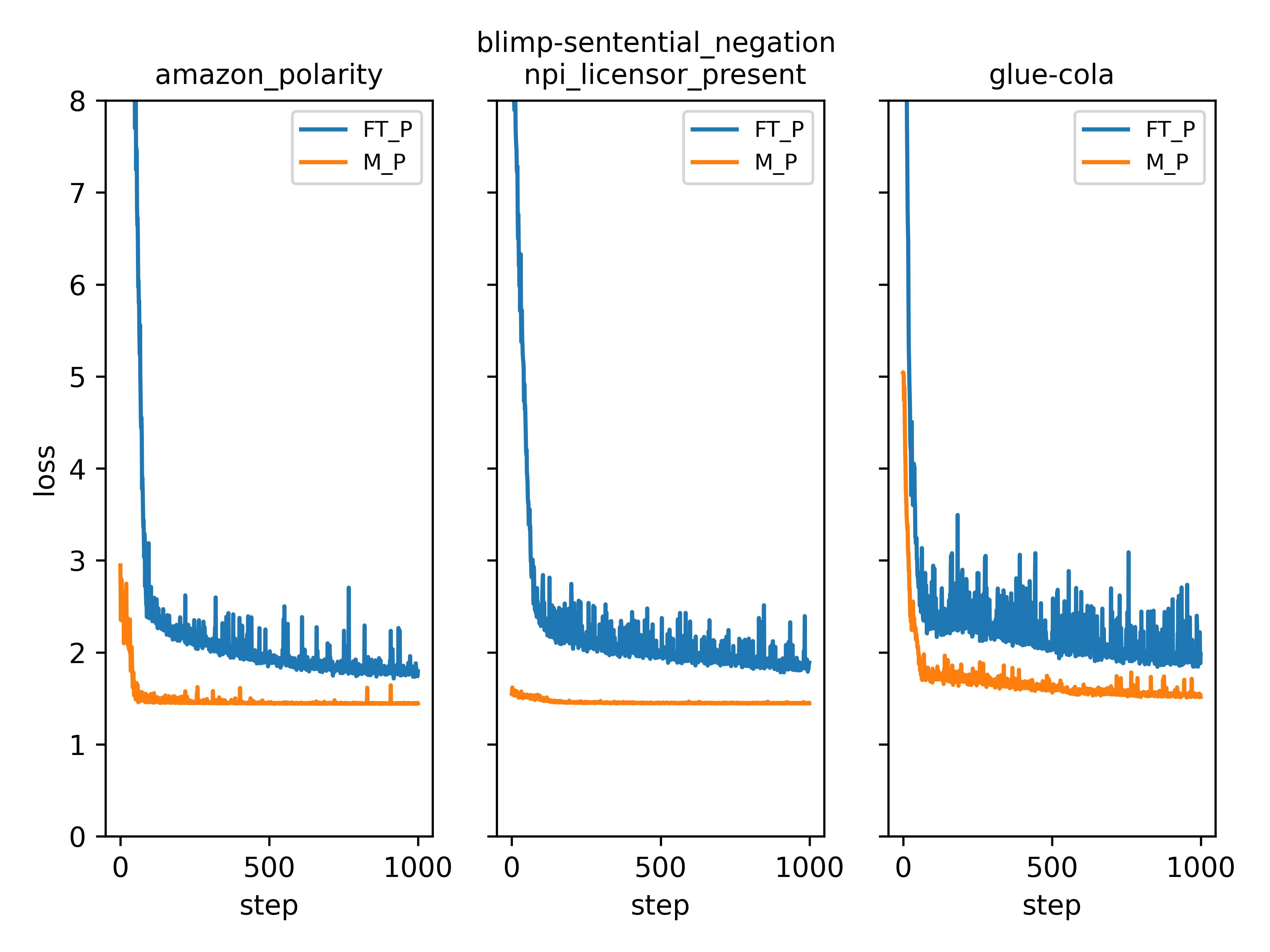}
        \caption{The loss curves of w \& w/o priming prompt in the fine-tuning stage.}
        \label{fig:train_loss}
    \end{figure}

    In this section, we emphasize the advantages of priming by examining training loss during downstream fine-tuning. We compare loss curves between models w/ or w/o priming (FT\_P and MTL M\_P, respectively) across three diverse tasks from the target task training set. These tasks, unseen by prompts/adapters, feature distinct evaluation metrics. Figure \ref{fig:train_loss} presents the results, where the blue curves represent FT\_P (w/o priming) and the orange curves represent MTL M\_P (w/ priming). The primed model not only converges faster but also achieves a superior final level of loss. Furthermore, the orange curves exhibit steadier convergence compared to the fluctuating and glitch-prone blue curves. These findings underscore the benefits of priming for prompts/adapters, facilitating rapid adaptation to various target tasks.
        
        
\section{Conclusion}
In this paper, we systematically analyze priming PEFT within a comprehensive framework. Our framework not only incorporates existing priming approaches but also explores previously uncharted strategies. Our experimental results demonstrate that the majority of priming strategies enhance the performance of PE methods. Notably, "Priming PLM only" emerges as the top-performing strategy when used in conjunction with multi-task learning. Crucially, our study provides concrete evidence that priming significantly facilitates the convergence of fine-tuning prompts/adapters on unseen tasks, underscoring the efficacy of priming.

\section{Limitation}
    We provide a systematic analysis of different priming strategies on PE methods and successfully improve the few-shot performance on diverse downstream tasks. However, there are some limitations to our work. Though we empirically show that MTL outperforms meta-learning, there are no further explanations for it. Besides, a small proportion of the priming strategies lead to a performance drop, but the actual reason remains unexplained. In addition, all the experiments are conducted on the pre-trained BART-base model. Extra experiments on other large language models may strengthen the results.


\section*{Acknowledgement}

We thank National Center for High-performance Computing (NCHC) of National Applied Research Laboratories (NARLabs) in Taiwan for providing computational and storage resources.

\bibliography{anthology,custom}
\bibliographystyle{acl_natbib}
\clearpage
\end{document}